\pdfoutput=1

\documentclass[11pt]{article}

\usepackage[final]{acl}

\usepackage{times}
\usepackage{latexsym}

\usepackage[T1]{fontenc}

\usepackage[utf8]{inputenc}

\usepackage{microtype}

\usepackage{inconsolata}

\usepackage{graphicx}

\usepackage{multirow}
\usepackage{caption}
\usepackage{tikz}
\usepackage{amsmath}
\usepackage{amssymb}
\usepackage{mathtools}
\usepackage{amsthm}

\newcommand{\nostart}{NoStart}
\newcommand{\coldstart}{ColdStart}
\newcommand{\warmstart}{WarmStart}
\newcommand{\hierarchical}{Hierarchical}
\definecolor{cvprblue}{rgb}{0.21,0.49,0.74}
\usepackage{subfiles}
\usepackage{subcaption}
\usepackage{pgfplots}
\usepackage{pgfplotstable}
\usetikzlibrary{calc}
\usetikzlibrary{fit}
\usetikzlibrary{patterns}
\usetikzlibrary{decorations.pathreplacing}
\usetikzlibrary{arrows,calc,arrows.meta}
\usepackage{diagbox}
\usepackage{multirow}
\usepackage{makecell}
\usepackage{xurl}
\usepackage[most]{tcolorbox}
\usepackage{xcolor}
\usepackage{lipsum}
\usepackage{hhline}

\definecolor{GPT4ocolor}{HTML}{636efa}
\definecolor{qwencolor}{HTML}{EF553B}
\definecolor{llamacolor}{HTML}{00CC96}
\definecolor{llavacolor}{HTML}{6D6D6D}
\definecolor{midlavacolor}{HTML}{6439FF}
\definecolor{vilacolor}{HTML}{FCCD2A}
\definecolor{openflamingocolor}{HTML}{117554}
\definecolor{minigpt4color}{HTML}{0D92F4}
\definecolor{mistralcolor}{HTML}{77CDFF}
\definecolor{bunnycolor}{HTML}{C62E2E}

\definecolor{NoOptColor}{HTML}{264653}
\definecolor{RandomColor}{HTML}{f4a261}
\definecolor{GreedyColor}{HTML}{7081ff}
\definecolor{BlueColor}{HTML}{0081a7}

\usepackage[capitalize,noabbrev]{cleveref}

\title{\textsc{Cache-of-Thought}: Master-Apprentice Framework for Cost-Effective Vision Language Model Reasoning}

\author{%
 Mingyuan Wu$^{1}$\thanks{~contributed equally to this work}, Jize Jiang$^{1*}$,  Haozhen Zheng$^{1*}$, \\  
 \bf Meitang Li$^{2}$, Zhaoheng Li$^{1}$, Beitong Tian$^{1}$, Bo Chen$^{1}$, \\
 \bf  Yongjoo Park$^{1}$, Minjia Zhang$^{1}$, Chengxiang Zhai$^{1}$, Klara Nahrstedt$^{1}$\\
$^{1}$University of Illinois Urbana Champaign\\
$^{2}$University of Michigan Ann Arbor \\
\texttt{\{mw34, jizej2, haozhen3, klara\}@cs.illinois.edu}  \\
}

\NewDocumentCommand{\cheng}{ mO{} }{\textcolor{purple}{\textsuperscript{\textit{Cheng}}\textsf{\textbf{\small[#1]}}}}

\begin{document}
\maketitle
\begin{abstract}
Vision Language Models (VLMs) have achieved remarkable success in a wide range of vision applications of increasing complexity and scales, yet choosing the right VLM model size involves a trade-off between response quality and cost. While smaller VLMs are cheaper to run, they typically produce responses only marginally better than random guessing on benchmarks such as MMMU. 

In this paper, we propose \textit{Cache of Thought (CoT)}, a master–apprentice framework for collaborative inference between large and small VLMs. CoT manages high-quality query results from large VLMs (\textit{master}) in a cache, which are then selected via a novel multi-modal retrieval and in-context learning to aid the performance of small VLMs (\textit{apprentice}). We extensively evaluate CoT on various widely-recognized and challenging general reasoning benchmarks, and show that CoT increases overall reasoning performance by up to 7.7\% under the same budget, and specifically boosts the reasoning performance of apprentice VLMs by up to 36.6\%. Our code is available at \url{https://github.com/UIUC-MONET/Cache-of-Thoughts}.

\end{abstract}
\section{Introduction}
\label{sec:intro}


Recent Vision Language Models (VLMs) \citep{gpt4o, Sonnet, gemini} have shown tremendous promise in a wide range of real-world applications, such as autonomous driving \citep{ad,RAG-drive}, robotics \citep{rt2, manipulate, pgvlm2024}, personalized virtual assistants \citep{assistant}, search engines \citep{search} and recommendation \citep{rec}. However, the ever-growing size of these recent VLMs has made at-scale deployment and operation challenging due to high consumption of cloud computing resource, high latency, and expensive API calls. 

\begin{figure}[t]\captionsetup[subfigure]{font=footnotesize}
\pgfplotsset{scaled y ticks=false}
\centering
\begin{tikzpicture}
\begin{axis}[
    clip=false,
    xtick=data,
    width=82mm,
    height=55mm,
    ymin=0,
    ymax=100,
    ylabel style={yshift = -3ex},
    axis y line*=none,
    axis x line*=none,
    ytick={0, 20, 40, 60, 80, 100},
    ytick={0, 20, 40, 60, 80, 100},
    xlabel=MMMU Validation Score,
    xlabel style={yshift = 1ex},
    label style={font=\scriptsize},
    ylabel style={xshift=-0.5ex, font=\scriptsize, align=center},
    xmin = 20,
    xmax = 80,
    xtick = {20, 30, 40, 50, 60, 70, 80},
    xticklabels = {20, 30, 40, 50, 60, 70, 80},
    tick label style={font=\scriptsize},
    x tick label style={yshift=0.5ex},
    legend style={
        at={(0.4,1.1)},anchor=south west,column sep=2pt,
        draw=black,fill=white,
        inner ysep=0.1pt,
        /tikz/every even column/.append style={column sep=5pt},
        font=\footnotesize
    },
    legend cell align={left},
    legend columns=3,
    ylabel={Model size (Billion)},
    ymajorgrids,
    every axis plot/.append style={thick},
    scatter/classes={
        a={mark size = 2pt, GPT4ocolor},
        b={mark size = 2pt, qwencolor},
        c={mark size = 2pt, llamacolor},
        d={mark size = 2pt, llavacolor},
        e={mark size = 2pt, midlavacolor},
        f={mark size = 2pt, vilacolor},
        g={mark size = 2pt, openflamingocolor},
        h={mark size = 2pt, minigpt4color},
        i={mark size = 2pt, mistralcolor},
        j={mark size = 2pt, bunnycolor}
    },
]

\addplot[
        scatter, 
        only marks,
        scatter src=explicit symbolic,
        nodes near coords*={\annotvalue},
        node near coord style={anchor=west, font=\scriptsize},
        visualization depends on={value \thisrow{annotation} \as \annotvalue}
    ]
table[meta=label,col sep=comma] {
        x,y,label,annotation
        69.1,110,a,{}
        64.5,72,b,{Qwen2-VL}
        60.3,90,c,{Llama 3.2}
        56.8,72,d,{}
        48.1,34,e,{}
        55.2,40,f,{InternVL2}
        28.7,9,g,{}
        26.8,13,h,{}
        35.3,7,i,{}
        38.2,3,j,{}
    };
\draw[red, thick, densely dotted] (axis cs: 22.1, 0) -- (axis cs: 22.1, 100);
\node[anchor=south west, align=right] at (axis cs: 22.1, 80) {\small \textcolor{red}{Random choice}};

\draw[black, opacity=0.7] (axis cs: 26.8, 15) -- (axis cs: 26.8, 30);
\node[anchor=south west, align=right, inner sep = 0mm] at (axis cs: 25.8, 32) {\scriptsize {MiniGPT4-Vicuna}};

\draw[black, opacity=0.7] (axis cs: 28.7, 11) -- (axis cs: 28.7, 22);
\node[anchor=south west, align=right, inner sep = 0mm] at (axis cs: 27.7, 23) {\scriptsize {OpenFlamingo2}};

\draw[black, opacity=0.7] (axis cs: 35.3, 9) -- (axis cs: 35.3, 14);
\node[anchor=south west, align=right, inner sep = 0mm] at (axis cs: 34.3, 15) {\scriptsize {LLaVA-NEXT-Mistral}};

\draw[black, opacity=0.7] (axis cs: 38.2, 5) -- (axis cs: 38.2, 7);
\node[anchor=south west, align=right, inner sep = 0mm] at (axis cs: 37.2, 6) {\scriptsize {Bunny}};

\draw[black, opacity=0.7] (axis cs: 48.1, 36) -- (axis cs: 48.1, 48);
\node[anchor=south west, align=right, inner sep = 0mm] at (axis cs: 47.2, 49) {\scriptsize {LLaVA-NEXT}};

\draw[black, opacity=0.7] (axis cs: 56.8, 74) -- (axis cs: 56.8, 77);
\node[anchor=south west, align=right, inner sep = 0mm] at (axis cs: 55.8, 78) {\scriptsize {LLaVA-OneVision}};

\draw[black, opacity=0.7] (axis cs: 69.1, 108) -- (axis cs: 69.1, 104);
\node[anchor=north west, align=right, inner sep = 0mm] at (axis cs: 62.1, 103) {\scriptsize {GPT-4o \textbf{(size outlier)}}};

\end{axis}
\end{tikzpicture}
\caption{Model Performance Comparison on MMMU(Val) as of Nov. 30, 2024\citep{mmmu}.
}
\label{fig:experiment_subindex_scatterplot}
\end{figure}
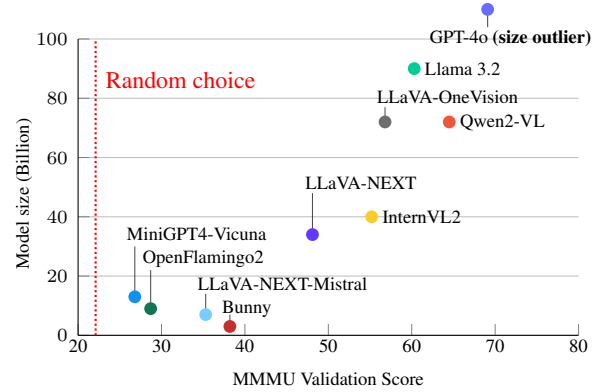

In response to this, there has been research focusing on developing smaller VLMs for on-device capabilities or cheap inference, such as MobileVLM-1.7B \citep{mobilevlm,mobilevlm2}, GPT-4o-mini \citep{gpt4omini} and Qwen-VL 7B \citep{Qwen2VL} etc. Unfortunately, one could not simply replace the larger VLM with a smaller one and expect the result to be within some practical tolerance, as the performance gap between large and small VLMs is still too \textit{huge}  (\Cref{fig:experiment_subindex_scatterplot}): compared to the impressive performance of large VLMs, some smaller VLMs offer only marginal improvements over random guessing on challenging reasoning benchmarks such as MMMU \citep{mmmu, uouo}.

In this paper, we seek a middle ground between smaller and larger models through collaboration between models of different sizes. To enhance on-the-fly performance of smaller models for a cost-effective reasoning system, we propose \textit{Cache of Thought (CoT)}, a master-apprentice framework that enables small (\textit{apprentice}) VLMs to generate responses of significant closer quality versus large (\textit{master}) VLMs via a novel design of dynamic \textit{Cache}. The Cache stores historical high-quality answer responses generated by master VLMs, which serves as a guidance for apprentice VLM query answering. Inspired by the principle of well-established case-based reasoning \citep{cbr}, CoT performs this guidance from master to apprentice via a specialized form of Retrieval-Augmented Generation (RAG): when CoT selects apprentice VLMs for question answering, it retrieves the most similar historical queries and responses in the cache to provide to the apprentice VLM as in-context examples. Through in-context learning, apprentice can benefit from the accumulated cached history of master, and become more capable of handling new queries. Notably, since the cache is allowed to grow, it is expected that the quality and relevancy of in-context samples should grow as well, leading to further improved capabilities of the apprentice VLM and the overall system. From the practicality perspective, CoT's in-context learning incurs negligible cost: it prepends the retrieved queries to the inference prompt of apprentice VLMs without introducing any extra training workloads, additional data annotations, nor recomputation expenses. CoT is also complementary to other VLM serving strategies and highly adaptable: as VLM routing and selection strategies mature, CoT can be easily integrated, continuing to offer significant benefits to smaller VLMs.

To the best of our knowledge, CoT is the first framework that demonstrates VLM in-context learning, combined with multimodal retrieval, can be applied effectively to real-world reasoning tasks with lengthy question prompts (e.g., MMMU \citep{mmmu}). In comparison, previous works explored VLM in-context learning under more constrained conditions, such as short textual image descriptions or questions of several words (e.g. "what is in the image") \citep{flamingo, manyshot}, purely visual contexts \citep{visualcontext}, or synthetic datasets with short, simple questions \citep{vl-icl}. These works often relied on straightforward top-$k$ image embeddings (or even random selection) for selecting in-context examples and utilized ground-truth (rather than model-generated) responses. Prior to this work, it was unclear which retrieval methods were most appropriate for dual-modality similarity, especially for complex VQA questions and in-context learning scenarios. CoT proposes and systematically evaluates various retrieval techniques (e.g., CLIP-based text+image embeddings, keyword extraction, keyword clustering), and maintains a dynamic cache of master VLM-generated responses, thereby expanding the scope and effectiveness of in-context learning for general, real-world VQA tasks without using ground truths.

In summary, we contribute (1) CoT, a general master-apprentice framework for multi-VLM inference with a cache, (2) an effective multi-modal retrieval and in-context learning approach that boosts apprentice VLM performance on the fly without training, and (3) extensive experiments demonstrating the effectiveness of CoT's multi-modal in-context learning and desirability of CoT's cost-response quality trade-off for general reasoning.

\begin{figure*}[]
	\centering
	\includegraphics[width=\textwidth]{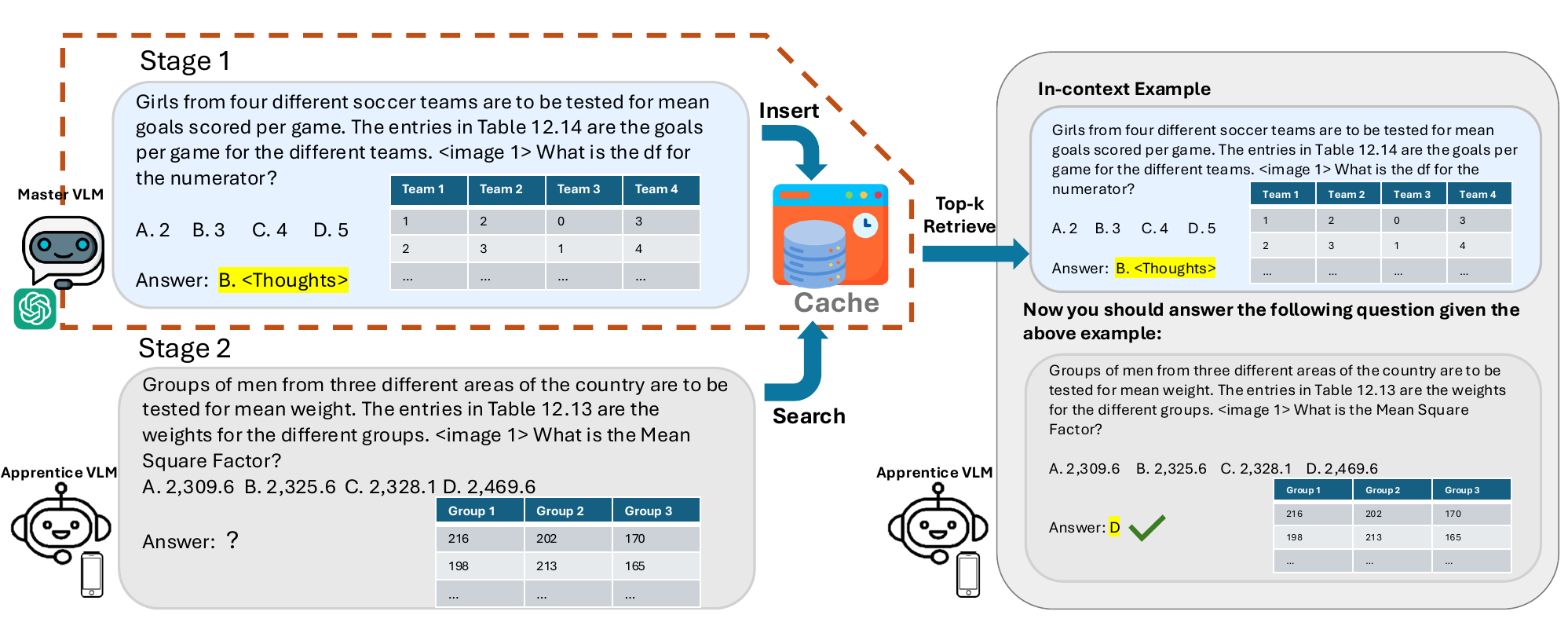}
	\caption{Apprentice VLM answers new query with help of past similar cases answered by the master VLM:\textit{ Multi-modal Retrieval and In-context learning}. Images and prompt examples cited from MMMU \citep{mmmu}}
	\label{fig:RAG}
\end{figure*}

\section{Related Work}
\label{sec:related}

\textbf{RAG and Multi-modal RAG}. RAG was originally proposed for language tasks \citep{RAG}. In RAG, a retriever is designed to extract relevant document chunks by similarity, and then these chunks are prepended to question prompts. RAG can reduce hallucinations, support knowledge-intensive tasks \citep{RAGsurvery}, and improve capability of small models \citep{ragsv}. Recent research has expanded RAG's capabilities beyond text. Multimodal RAG \citep{ragvl, finegrain, lin-etal-2024-preflmr, murag, RACM3} retrieves world knowledge from relevant multimodal documents and improves knowledge-seeking or fact-based VQA tasks \citep{FVQA, okvqa, info, evqa}. CoT differs from multi-modal RAG in that its cache is dynamic, and all stored responses are generated by a large master VLM rather than factual documents. 

\noindent \textbf{In-context Learning and Multi-modal In-context Learning}. In-context learning emerged with large auto-regressive models like GPT-3 \citep{gpt3}, which adapt to new tasks by observing a few in-context demonstrations. This success motivated research into VLMs. Flamingo \citep{flamingo} demonstrated significant gains in VQA by providing random demonstrations in context, while \citet{visualcontext} extended these benefits to tasks like image segmentation and classification. More recently, \citet{vl-icl} validated in-context learning for VLMs ranging from 4B to 70B parameters using synthetic VQA. CoT makes unique contribution in multi-modal in-context learning, with we detail in \cref{sec:incontext}.

\section{Cache-of-Thought Framework}
\label{sec:system_components}
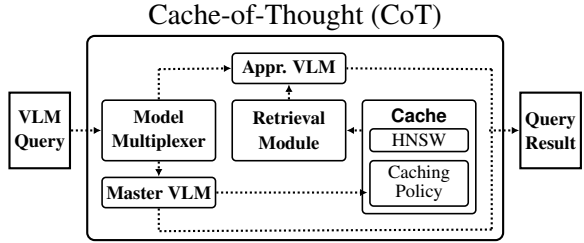
\begin{figure}[t]

\begin{subfigure}{\columnwidth}
\centering
\tikzset{
mylabel/.style={
    font=\footnotesize\sffamily,
    align=center,
},
mylabel2/.style={
    font=\footnotesize\sffamily,
    align=center,
},
mycomponent/.style={
    font=\footnotesize\sffamily\bfseries,
    semithick, rounded corners=0.5mm,
}
}

\begin{tikzpicture}[>={LaTeX[width=1mm,length=1mm]},->]

\node(shell) [draw=black, anchor=west, minimum width = 8mm, minimum height=10mm, thick]
at (0, 0)  {};
\node(shelltxt) [anchor=center, align = center, minimum height=3mm, inner sep = 0.5mm, font=\bfseries\actualfootnotesize]
at ($(shell.center) + (0, 0)$) {VLM \\ Query};

\node(elastic) [minimum height=27mm, minimum width=55mm,
  draw=black, anchor=west, thick, rounded corners=1mm]
at ($(shell.east) + (0.2, -0.1)$)  {};
\node(elastictxt) [anchor=south, minimum height=3mm, inner sep = 0mm]
at ($(elastic.north) + (0, 0.05)$) {Cache-of-Thought (CoT)};

\node(multiplexer) [draw=black, fill = white, anchor=west, minimum width = 15mm, minimum height=8mm, mycomponent]
at ($(elastic.west) + (0.2, 0.1)$)  {};

\node(multiplexertxt) [anchor=center, align = center, minimum height=3mm, inner sep = 0.5mm]
at ($(multiplexer) + (0, 0)$) {\actualfootnotesize \textbf{Model} \\[-0.4em] \actualfootnotesize\textbf{Multiplexer}};

\node(largevlm) [draw=black, fill = white, anchor=north, minimum width = 15mm, minimum height=4mm, mycomponent]
at ($(multiplexer.south) + (0, -0.2)$)  {};

\node(largevlmtxt) [anchor=north, align = center, minimum height=3mm, inner sep = 0.5mm]
at ($(largevlm.north) + (0, -0.05)$) {\actualfootnotesize \textbf{Master VLM}};

\node(retriever) [draw=black, fill = white, anchor=west, minimum width = 15mm, minimum height=8mm, mycomponent]
at ($(multiplexer.east) + (0.2, 0)$)  {};

\node(retrievertxt) [anchor=center, align = center, minimum height=3mm, inner sep = 0.5mm]
at ($(retriever) + (0, 0)$) {\actualfootnotesize \textbf{Retrieval} \\[-0.4em] \actualfootnotesize \textbf{Module}};

\node(smallvlm) [draw=black, fill = white, anchor=south, minimum width = 15mm, minimum height=4mm, mycomponent]
at ($(retriever.north) + (0, 0.2)$)  {};

\node(smallvlmtxt) [anchor=north, align = center, minimum height=3mm, inner sep = 0.5mm, ]
at ($(smallvlm.north) + (0, -0.05)$) {\actualfootnotesize \textbf{Appr. VLM}};

\node(cache) [draw=black, fill = white, anchor=north west, minimum width = 15mm, minimum height=15mm, mycomponent]
at ($(retriever.north east) + (0.2, 0)$)  {};

\node(cachetxt) [anchor=north, align = center, minimum height=3mm, inner sep = 0.5mm, mylabel]
at ($(cache.north) + (0, -0.05)$) {\actualfootnotesize \textbf{Cache}};

\node(evicter) [draw=black, fill = white, anchor=south, minimum width = 13mm, minimum height=6mm, mycomponent]
at ($(cache.south) + (0, 0.1)$)  {};

\node(evictertxt) [anchor=center, align = center, minimum height=3mm, inner sep = 0.5mm]
at ($(evicter) + (0, 0)$) {\scriptsize Caching \\[-0.6em]\scriptsize Policy};

\node(hnsw) [draw=black, fill = white, anchor=south, minimum width = 13mm, minimum height=3mm, mycomponent]
at ($(evicter.north) + (0, 0.1)$)  {};

\node(hnswtxt) [anchor=center, align = center, minimum height=3mm, inner sep = 0.5mm]
at ($(hnsw) + (0, 0)$) {\scriptsize HNSW};

\node(results) [draw=black, anchor=west, minimum width = 8mm, minimum height=10mm, thick]
at ($(elastic.east) + (0.2, 0.1)$)  {};
\node(resultstxt) [anchor=center, align = center, minimum height=3mm, inner sep = 0.5mm, font=\bfseries\actualfootnotesize]
at ($(results.center) + (0, 0)$) {Query \\ Result};

\draw[->, thick, densely dotted] 
($(shell.east)$) --
($(multiplexer.west)$);

\draw[->, thick, densely dotted] 
($(multiplexer.south)$) --
($(largevlm.north)$);

\draw[->, thick, densely dotted] 
($(largevlm.east)$) --
($(largevlm.east) + (2.0, 0)$);

\draw[->, thick, densely dotted] 
($(retriever.east) + (0.2, 0)$) --
($(retriever.east)$);

\draw[->, thick, densely dotted] 
($(retriever.north)$) --
($(smallvlm.south)$);

\draw[-, thick, densely dotted] 
($(multiplexer.north)$) --
($(multiplexer.north) + (0, 0.4)$);

\draw[->, thick, densely dotted] 
($(multiplexer.north) + (0, 0.4)$) --
($(smallvlm.west)$);

\draw[-, thick, densely dotted] 
($(largevlm.south)$) --
($(largevlm.south) + (0, -0.3)$);

\draw[-, thick, densely dotted] 
($(cache.south east) + (0.2, -0.2)$) --
($(largevlm.south) + (0, -0.3)$);

\draw[-, thick, densely dotted] 
($(cache.north east) + (0.2, 0.4)$) --
($(smallvlm.east)$);

\draw[-, thick, densely dotted] 
($(cache.north east) + (0.2, 0.4)$) --
($(cache.south east) + (0.2, -0.2)$);

\draw[->, thick, densely dotted] 
($(results.west) + (-0.4, 0)$) --
($(results.west)$);

\end{tikzpicture}
\end{subfigure}
\caption{The Cache-of-Thought (CoT) framework.}
\label{fig:system_overview}
\end{figure}
CoT (\cref{fig:system_overview}) is a query serving framework for VLM queries that interleaves large (\textit{master}) and small (\textit{apprentice})  VLM calls for significant performance boost and cost savings: query results from master VLM calls are cached, which are then used to enhance apprentice VLM calls via multi-modal retrieval and in-context learning.

\noindent \textbf{Master VLM}. CoT's master VLM (often with several hundred billion parameters and serves with per-token charge), under the assumption that it produces high-quality answers, acts as the generator of the QA-pairs stored in the cache.

\noindent \textbf{Apprentice VLM}. CoT's apprentice VLM (often with less than 7 billion parameters) is used to answer queries via augmentation with various in-context examples fetched from the cache.

\noindent \textbf{Model Multiplexer}. The Model Multiplexer routes incoming VLM queries by choosing one of two serving methods: use the master VLM to directly answer the query, or use the apprentice VLM to answer the query augmented with in-context examples fetched from the Cache (described shortly). The multiplexer balances cost and quality by controlling how often CoT calls the master and apprentice VLMs: frequently calling the master VLM increases the rate of populating examples into the cache (hence more effective in-context learning) but incurs a higher cost, and vice versa.

\noindent \textbf{Cache.} CoT's cache stores high-quality QA pairs from the master VLM. When the multiplexer routes a new query to the master VLM, the question asked, image prompt, and resulting answer (QA-pair) is stored into the cache, which then computes its dual-modality embedding from the question and image and inserts it into a HNSW index \citep{HNSW} to facilitate accurate retrieval as an in-context example. 
CoT's cache can also be pre-loaded with QA-pairs to efficiently warm-start on query serving, and can incorporate eviction policies---both common (e.g., LRU or LFU) or specialized, workload-aware~\cite{li2023sc} for more resource-constrained settings.\footnote{We defer concrete explorations on CoT's eviction policy to future work when real-world dual modality user query benchmarks are available.}

\noindent \textbf{Retrieval Module}. The Retrieval Module retrieves relevant in-context examples (i.e., QA-pairs) from the Cache to augment queries routed to the apprentice VLM by the Multiplexer. It performs an efficient ANN vector search with the dual-modality embedding of the incoming query in the Cache's HNSW index, retrieving examples similar to the query in terms of both the question asked and image prompt.
\section{Methodology}
This section describes CoT's methodology. \Cref{sec:method_preliminary} overviews CoT's formulation, \cref{sec:incontext} describes CoT's multi-modal in-context learning, and \cref{sec:retrieval} covers how CoT retrieves relevant examples for in-context learning.
\subsection{Formulation}
\label{sec:method_preliminary}

This section formulates CoT's notations. Without loss of generality, CoT deploys two VLM instances with frozen weights, the master VLM $f$ and apprentice VLM $g$, where deployment cost of $g$ is significantly lower than $F$. CoT's workload is a set of (finite) queries $\mathcal{Q} \subset \mathbb{R}^d$: At each round $t$, the system receives a query from the workload $(x_t, q_t)\in \mathcal{Q}$ consisting of the visual input $x_t$ and the textual question $q_t$ (about the content of $x_t$ e.g., 'how many ducks are there in the picture?').

For each query $(x_t, q_t)$ from round $t$, CoT's model multiplexer $J$ determines which VLM to use (the master $f$ or apprentice $g$) with the multiplexer function: $J: \mathcal{Q} \mapsto\{f, g\}$, which routes the query according to $J((x_t, q_t))$. Then, the selected VLM will serve the query by generating an answer from an answer space $\mathcal{A}$ with the respective function $f: \mathcal{Q} \mapsto \mathcal{A}$ or $g: \mathcal{Q} \mapsto \mathcal{A}$, e.g., $g((x_t, q_t)) = a_t \in \mathcal{A}$. $x_t$, $q_t$, and the answer from the master/apprentice VLM $a_t$ will then comprise the \textit{QA-pair} $P_t = \{(x_i, q_i), a_i\}$.

CoT's cache, denoted as $\mathcal{L}_t \subset \mathcal{Q}$, will store $\left|\mathcal{L}_t\right| \leq L$ QA-pairs $P_t$ generated with $g$, i.e., $P_t$ s.t. $J((x_t, q_t)) = g$, where $L$ is the cache capacity. CoT inserts into $\mathcal{L}_t$ whenever $g$ is invoked; however, due to the bounded size $L$, insertions performed when $\left|\mathcal{L}_t\right| = L$ will trigger evictions. Initially, $\mathcal{L}_0$ can either be empty (cold start) or a prepopulated set of QA-pairs (warm start) from the master VLM $g$. Hence, the cache $\mathcal{L}_t$ serves as an external knowledge base to support the apprentice VLM $f$ during inference via multi-modal retrieval and in-context learning (\cref{sec:incontext}). 

For retrieval from the Cache $\mathcal{L}_t$, CoT uses a \textit{retrieval mechanism} $R$ to retrieve the top-k relevant QA-pairs $\mathcal{L}_k$ from the cache: $R(\mathcal{L}_{t_i}, (x_{t_j}, q_{t_j})) \mapsto \mathcal{L}_k \subseteq \mathcal{L}_{t_i}, j > i$, which we describe in \cref{sec:retrieval}.
\subsection{Multi-modal In-context learning}
\label{sec:incontext}
CoT performs multi-modal in-context learning by transferring knowledge from the cache to the apprentice VLM. Given the pre-trained apprentice VLM $f$ with frozen parameters $\theta$ and a query $(x_t, q_t)\in \mathcal{Q}$, CoT retrieves a set of QA-pairs from the cache $M = \{(x_i, q_i), a_i\}$ from $\mathcal{L}_t$ used to aid inference. The apprentice VLM $f$ then generates an answer $a_t$ in one forward pass: $a_t = f_\theta\left( (x_t, q_t), M\right)$.

\noindent \textbf{Flexibility for Dynamic Systems}.
Compared to finetuning apprentice VLM instances with Lora using cached content \citep{lora}, CoT's in-context learning offers several advantages in dynamic systems: (1) CoT pre-computes QA-pairs and stores them in the cache for reuse, incurring no extra data annotation and/or recomputation expenses during retrieval; hence, CoT's performance will steadily improve with minimal costs as it answers more high-quality queries with the master VLM. (2) CoT's self-improvement via in-context learning does not require weight updates for its VLMs, hence there are no extra training overheads or delays (e.g., waiting for model convergence).

\noindent \textbf{Prompt Construction}. For each query routed to the apprentice VLM, CoT performs in-context learning by retrieving one or more demonstration examples to prepend to the inference prompt. CoT utilizes special tokens for prompt formatting, interleaving support QA-pairs in the form of (image,text): "\{support image\}, Question: \{support question\}, Answer: \{\textit{ground truth answer}\}, \{query image\}, Question: \{query question\}, Answer:" In CoT, \{\textit{ground truth answer}\} is replaced by stored VLM responses from the cache $\mathcal{L}_t$. Beyond this core prompt structure, additional components---such as n-shots and system prompts required by the VLM’s input format---are incorporated in the final prompt. For effective prompt construction, CoT explicitly instructs the master VLM to "include reasoning steps" when using it to serve queries to obtain complete responses (to prepend to prompts sent to the apprentice VLM) rather than just short answers or multiple-choice selections. This allows the apprentice VLM to better leverage the master VLM's reasoning capabilities. Full prompts details can be found in the \cref{sec:appendix}.

\subsection{Multi-modal Retrieval}
\label{sec:retrieval}
In this section, we describe CoT's multimodal retrieval of in-context examples from its Cache, on which performance on downstream tasks is highly sensitive to \citep{visualcontext}. Much to our surprise, existing works on VLM in-context learning present ungrounded design choices without further experiments in the following aspects:

\noindent \textbf{Choice of Retrieval}. Recent works \citep{vl-icl, manyshot} randomly select in-context examples for all test instances. Flamingo \citep{flamingo} attempts retrieval-based in-context example selection, which retrieves top-k similar samples using frozen pre-trained vision encoder embeddings from CLIP \citep{CLIP}. While Flamingo verifies that its image-driven retrieval is better than random selection in test cases with short, few-word questions, it remains unclear whether this method can be extend to general VQA systems where questions can be long and complicated, as seen in benchmarks like MMMU \citep{mmmu}.

\noindent \textbf{Choice of Support Set}. All of the aforementioned studies randomly partition the dataset into a support set and test set offline. However, in-context performance can heavily depend on the distribution of support examples. A setting where the support set is dynamic (i.e., continuously growing as in CoT's case) has not been explored.

\noindent \textbf{Assumption of Ground-Truth Availability}. Previous studies often use in-context examples with human-annotated ground truth answers, which can be inpractical for real-world, online data streams.

In contrast, CoT targets a more challenging and general VQA setting, and makes significant contributions to multi-modal retrieval for in-context learning in terms of the aforementioned 3 aspects: (1) CoT proposes a retrieval method that leverages both CLIP image and text embeddings, supplemented by keyword-based techniques for long or complex queries. CoT also introduces a hierarchical keyword clustering approach tailored to the query distribution (\cref{sec:retrieval_method}). (2) CoT's in-context example is performed over a dynamically updating (as opposed to static in previous works) knowledge base $\mathcal{L}_t$, where new items (that align more closely with recent queries) are continuously added to the cache. (3) CoT directly utilizes responses from the master VLM as in-context examples rather than the human-labeled ground truth answers strongly assumed by all prior works. This allows CoT to incorporate more diverse examples as long as they have been seen by the master VLM, rather than relying on human annotators.

\subsubsection{Dual-Modality Dense Retrieval}
\label{sec:retrieval_method}
We begin with a simple dense-retrieval method that uses a unified embedding vector to identify the top-\emph{k} matches. A natural starting point is CLIP \citep{CLIP}, which has proven effective in multimodal retrieval \citep{RACM3, inquire} and is pretrained to align image and text embeddings on internet-scale (image, text) pairs. Recognizing CLIP as a well aligned dual-modality representation, we use its image and text encoders to separately embed each image and text, then average these embeddings to form a unified vector that is stored in our cache index and use for query embeddings. 

However, CLIP embeddings also have limitations. They are pretrained primarily on short textual descriptions of images (e.g., “a photo of a cat”) and support a relatively small text context window—often fewer than 20 tokens effectively \citep{longclip}. This limitation makes CLIP less ideal for lengthy, sentence-like questions. To mitigate this issue, we break down long questions before encoding them with CLIP. Specifically, we employ a small auxiliary LLM, denoted as $Extractor$ to extract keywords from the long sentence-like query. Details of how we prompt the $Extractor$ can be found in the Appendix.

In some cases, the question text alone may not be sufficiently informative for retrieval (e.g. a question like "what is in the image"); Fortunately, CoT's cache also stores responses from the master VLM, allowing us to extract text embeddings from these richer responses instead of relying solely on the original question. For new queries, we can even make a single pass with the apprentice VLM to generate a text answer response, extract keywords from that response, then feed them into CLIP text encoder to enhance retrieval. Because these auxiliary or apprentice LLM calls are much cheaper than invoking the master VLM, we can use them to achieve significantly better retrieval results without noticeably increasing inference cost. This strategy, which combines embeddings from both the text response and the corresponding image, significantly improves the robustness of our dense-retrieval process, particularly in VQA scenarios where the question alone lacks sufficient context.

\subsubsection{Multi-stage Hashtag Retrieval}
In real-world scenarios, the streaming data distribution $\mathcal{Q}$ is highly complex and diverse, varying across different contexts. When the cache is large, directly performing dense retrieval from the entire knowledge base $\mathcal{L}_t$ may fail to locate the relevant contexts due to the limited expressiveness of CLIP embeddings even if they are carefully designed (\cref{sec:retrieval_method}). A straightforward approach to improve retrieval accuracy is to restrict retrieval to a specific knowledge domain. However, without a predefined distribution for $\mathcal{Q}$, categorizing streaming data under fixed domains is often difficult.

To enable more granular unsupervised retrieval, we design a two-stage hashtag tree where Level 1 hashtags $\mathcal{H}_1$ capture high-level, task-oriented representations, and Level 2 hashtags $\mathcal{H}_2$ provide finer, concept-specific details. Each data entry in $\mathcal{L}_t$ is assigned at least one Level 1 hashtag and one Level 2 hashtag. We use $h_{i,j}$ to denote the $j^{th}$ hashtag at Level $i$, where $\mathcal{H}_i = \{ h_{i,j} \}$, $i = 1, 2$.

Initially, both the hashtag tree and the cache $\mathcal{L}_t$ are empty. The master VLM g is tasked with inferring $a_t$ given each new-coming data $(x_t, q_t)$ during the cold start stage, after which the Level 2 hashtag $h_{2,t}$ is obtained as: $h_{2,t} = CLIPTextEncoder(Extractor(a_t)) $.
Once $|\mathcal{H}_2| > \tau$ or the cold start ends, we use K-Means to cluster $\mathcal{H}_2$ into $K$ groups $\{C_k\}_{k=1}^{K}$, where each cluster is assigned a Level 1 hashtag $h_{1,k}$, computed as the mean embedding of its corresponding Level 2 hashtags. 

Given new data $t'$, when the master VLM $g$ is selected for inference, the generated answer is stored in $\mathcal{L}_t$ and assigned to the closest Level 1 hashtag $h_{1,j^*}$ based on Euclidean distance. The Level 1 hashtag is then updated as the weighted average of its current Level 2 hashtag children. Otherwise, if the apprentice VLM $f$ is selected to infer on $t'$, $\mathcal{R}$ first retrieves all examples from $\mathcal{L}_t$ that shares the same Level 1 hashtag as $d_t$. It then selects top-$k$ results as the $k$-shot in-context examples for $f$.\footnote{CoT currently implements this with hnswlib's~\cite{hnswlib} built-in filtered search mechanism; we defer incorporating dedicated filtered vector search indexes (e.g., ~\cite{li2025sieve}) into CoT to future work.}
\section{Experiment}
\label{sec:experiment}
This section describes our experimental evaluation of CoT. We design our experiments as follows: (1) We evaluate the performance gains on general VQA tasks achieved by CoT's applying of in-context learning to apprentice VLMs via multi-modal retrieval.
(2) We evaluate the sensitivity of CoT to different system configurations such as cache size and choice of retrieval methods.
(3) We evaluate CoT in both static and dynamic configurations (explained shortly). 
(4) We study CoT's trade-offs between performance and cost advantage (metric defined by \citep{hybridllm}) by adjusting the ratio of master and apprentice VLM usage.

Specifically, for (3), in the static configuration, We pre-construct $\mathcal{L}_t$ using offline data, which remains unchanged in each round $t$. We also fix the model multiplexer $J$ to only select the apprentice VLM. These settings allow us to evaluate how much the apprentice VLM potentially benefits from CoT under a static cache. In the dynamic configuration, CoT processes a continuous data stream and uses a dynamic cache $\mathcal{L}_t$ (explained shortly) which may start either empty or containing offline data. The model multiplexer randomly selects between the apprentice and master VLM according to a specified probability (e.g., a 10\% apprentice VLM rate means that it is selected for query serving 10\% of the time). After each round, CoT inserts into the cache with the new QA-pair (when the master VLM is used) and may perform eviction (when the cache is at capacity) to maintain cache effectiveness for downstream tasks. This setup allows us to evaluate both CoT's performance-cost trade-offs and capability to adapt over time.

\subsection{Experiment Datasets}

\textbf{MMMU} \citep{mmmu} is a widely recognized and challenging multi-disciplinary VLM benchmark that organizes subjects into a hierarchical structure. It evaluates VLMs on both breadth and depth of subject knowledge, as well as on their expert-level reasoning and understanding. MMMU covers complex VQA tasks across six major fields, ranging from art to science and businesses. The dataset consists of 11.5K questions, divided into development (\textit{dev}), validation (\textit{val}), and test sets (the test set does not include ground truth answers). To ensure fair evaluation of in-context learning, we filter out instances containing multiple images, as most off-the-shelf VLMs are less capable of handling multi-image questions, which could interfere with their in-context learning capabilities. After filtering, the sizes of the dev, val and test set are 146, 857, 9702 respectively.

\textbf{VL-ICL} \citep{vl-icl} is a pioneering synthetic benchmark designed to measure the broader capabilities and limitations of multi-modal in-context learning. We employ this dataset to examine whether CoT retains the benefits of in-context learning even when only responses generated by the master VLM instead of ground-truth annotations are used. We specifically adopt the TextOCR and Clevr subtasks, as the remaining subtasks lack meaningful question prompts in general language settings. The data set is divided into two subsets of 200 and 800 samples, which we refer to as dev and val, respectively, in this paper.

\subsection{Models}
We choose between the open-source Qwen-VL-2 7B \citep{Qwen2VL} model as a representative of recent VLMs, and open-flamingo 3B and 9B \citep{flamingo} (from an earlier generation of VLMs) as CoT's apprentice VLM in various experiments, to evaluate whether CoT can benefit older VLMs and newer, well-instruction tuned VLMs alike. We run all apprentice models with 4 A100 GPUs of 40GB. We use GPT4-o \citep{gpt4o} as CoT's master VLM. 
\begin{table}[t]
\caption{Qwen7B MMMU Score vs. Retrieval Strategy}
\label{sample-table}
\begin{center}
\realfootnotesize
\begin{tabular}{|c|c|c||c|c|}
\hline
\multicolumn{3}{|c||}{Method} & \makecell{Cache Val\\Test Dev} & \makecell{Cache Dev\\Test Val}\\
\hline
\multicolumn{3}{|c||}{No in-context} & 24.66 & 34.42\\
\hline
\multicolumn{3}{|c||}{Hierarchical Cluster based} & \textbf{39.04} & \textbf{39.44}\\
\hline
\multicolumn{3}{|c||}{GPT-4o} & 58.90 & 64.87 \\
\hline
Cache & Query & Cache size & & \\ 
\hhline{|=|=|=||~|~|}
Image & \makecell{Image+\\ Text} & Full &35.62&39.20\\
\hline
Image & \makecell{Image+\\ Text} & Half & 34.25& 38.74\\
\hline
\makecell{Image+\\ Text}  & \makecell{Image+\\ Text} & Full & 34.90& 38.39\\
\hline
Image  & Image & Full & 37.67& 37.92\\

\hline
\end{tabular}
\end{center}
\label{tab:Qwen}
\end{table}

\begin{table}[t]
\centering
\setlength\tabcolsep{2pt} 
\caption{OpenFlamingo w. Cache Val Test Dev Setting}
\actualfootnotesize
\begin{tabular}{ |c||c|c|c|c|c||c|c| }
 \hline
Model & \multicolumn{5}{|c||}{\makecell{OpenFlamingo-3B\\ (CLIP ViT-L/14, MPT-1B-Dolly)}} & \multicolumn{2}{|c|}{OpenFlamingo-9B }\\
 \hline
Cache size & N/A &  \multicolumn{2}{|c|}{Full Cache Size} & \multicolumn{2}{|c||}{Half Cache Size}& \multicolumn{2}{|c|}{Full Cache Size}\\
 \hline
 \diagbox[width=13mm]{Dataset}{N-shot} & 0-shot & 1-shot & 2-shot & 1-shot & 2-shot & 0-shot & 1-shot\\
 \hline
mmmu & 15.75 & 17.12 & & 17.12 &  &20.55 &25.34 \\
 \hline
clevr & 7.50 & 20.50 & 28.00  & 22.50 & 26.50& \multicolumn{2}{|c|}{\multirow{2}{*}{}}\\
 \cline{0-5}
textocr & 0.00 & 6.50 & 5.50 & 6.50 & 5.50 & \multicolumn{2}{|c|}{}\\
 \hline
\end{tabular}
\vfill
\label{tab:Flamingo1}
\end{table}
\begin{table}[t]
\centering
\setlength\tabcolsep{2pt} 
\caption{OpenFlamingo w. Cache Dev Test Val Setting}
\actualfootnotesize
\begin{tabular}{ |c||c|c|c|c|c||c|c| }
 \hline
Model & \multicolumn{5}{|c||}{\makecell{OpenFlamingo-3B\\ (CLIP ViT-L/14, MPT-1B-Dolly)}} & \multicolumn{2}{|c|}{OpenFlamingo-9B }\\
 \hline
Cache size & N/A &  \multicolumn{2}{|c|}{Full Cache Size} & \multicolumn{2}{|c||}{Half Cache Size}& \multicolumn{2}{|c|}{Full Cache Size}\\
\hline
 \diagbox[width=13mm]{Dataset}{N-shot} & 0-shot & 1-shot & 2-shot & 1-shot & 2-shot & 0-shot & 1-shot\\
 \hline
mmmu & 22.75 & 23.92 &  & 24.04 & & 24.97 & 26.25 \\
 \hline
clevr & 6.88 & 21.50 & 21.75 & 19.38 & 19.38 & \multicolumn{2}{|c|}{\multirow{2}{*}{}} \\
  \cline{0-5}
textocr & 0.00 & 5.00 & 3.38 & 4.63 & 2.13 & \multicolumn{2}{|c|}{} \\
 \hline
\end{tabular}
\label{tab:Flamingo2}
\end{table}

\subsection{Evaluation}
We use accuracy as the evaluation metric. We note that the main purpose of our evaluations is to present a holistic picture of how CoT would work for end users, thus the absolute performance of the models are not the priority. For both multiple-choice answers in MMMU and open responses in VL-ICL, we allow the VLMs to produce intermediate steps before generating the final answer. We prompt the VLMs to produce the final answer in a specific format, and use a rule-based parser to determine the VLMs’ choice. More specifically, for multipe-choice answers, we extract the VLMs' choice, and for open responses, we check whether there exists an exact match of the label in the VLMs’ response. If the parser fails to determine the VLMs’ choice, we assign a \texttt{None} value instead, since in reality, an randomly generated choice is misleading if not detrimental.

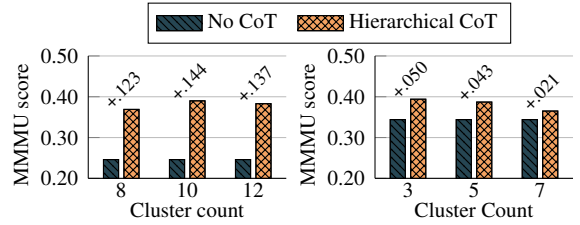
\begin{figure}[t]\captionsetup[subfigure]{font=footnotesize}
\pgfplotsset{scaled y ticks=false}
\centering
\begin{subfigure}[b]{0.48\linewidth}
\centering
\begin{tikzpicture}

\pgfplotstableread[col sep=comma,]{
name
8
10
12
}\datatable

\begin{axis}[
    ybar,
    clip=false,
    xtick={1, 2, 3, 4},
    xticklabels from table={\datatable}{name},
                 x tick label style={anchor=center, yshift = 0ex, font=\realfootnotesize},
    xtick style ={draw=none},
    xlabel style={yshift = 2ex, font=\realfootnotesize},
    ylabel style={yshift = -2ex, xshift = -0.6ex, font=\realfootnotesize},
    width=42mm,
    height=32mm,
    bar width=2.0mm,
    ymin=0.20,
    ymax=0.50,
    axis y line*=none,
    axis x line*=none,
    ytick={0.20, 0.30, 0.40, 0.50},
    yticklabels={0.20, 0.30, 0.40, 0.50},
    xmin=0.5,
    xmax = 3.5,
    ymajorgrids,
    tick label style={font=\realfootnotesize},
    legend style={
        font=\realfootnotesize,
        /tikz/every even column/.append style={column sep=0.2cm},
        legend columns = 3,
        at={(0.3,1.1)},
        anchor=south west,
    },
    xlabel={Cluster count},
    ylabel={MMMU score},
    area legend
    ]
    
\addplot [black,fill=NoOptColor, postaction={
        pattern=north west lines
    }] table[x=x,y=y] {
x y
1 0.246
2 0.246
3 0.246
};
\addlegendentry[]{No CoT};

\addplot [black,fill=RandomColor,postaction={
        pattern=crosshatch
    }] table[x=x,y=y] {
x y
1 0.369
2 0.390
3 0.383
    };
\addlegendentry[]{Hierarchical CoT};

\node[anchor=south, align=left, rotate=45] at (axis cs: 1.2, 0.389) {\scriptsize +.123};
\node[anchor=south, align=left, rotate=45] at (axis cs: 2.2, 0.410) {\scriptsize +.144};
\node[anchor=south, align=left, rotate=45] at (axis cs: 3.2, 0.403) {\scriptsize +.137};

\end{axis}
    
\end{tikzpicture}
\vspace{-6mm}
\caption{Cache Val Test Dev}
\end{subfigure}
\begin{subfigure}[b]{0.48\linewidth}
\centering
\begin{tikzpicture}

\pgfplotstableread[col sep=comma,]{
name
3
5
7
}\datatable

\begin{axis}[
    ybar,
    clip=false,
    xtick={1, 2, 3, 4},
    xticklabels from table={\datatable}{name},
                 x tick label style={anchor=center, yshift = 0ex, font=\realfootnotesize},
    xtick style ={draw=none},
    xlabel style={yshift = 2ex, font=\realfootnotesize},
    ylabel style={yshift = -2ex, xshift = -0.6ex, font=\realfootnotesize},
    width=42mm,
    height=32mm,
    bar width=2.0mm,
    ymin=0.20,
    ymax=0.50,
    axis y line*=none,
    axis x line*=none,
    ytick={0.20, 0.30, 0.40, 0.50},
    yticklabels={0.20, 0.30, 0.40, 0.50},
    xmin=0.5,
    xmax = 3.5,
    ymajorgrids,
    tick label style={font=\realfootnotesize},
    legend style={
        font=\realfootnotesize,
        /tikz/every even column/.append style={column sep=0.2cm},
        legend columns = 3,
        inner ysep=0.5pt,
        at={(0.4,1.3)},
        anchor=south west,
    },
    xlabel={Cluster Count},
    ylabel={MMMU score},
    area legend
    ]
    
\addplot [black,fill=NoOptColor, postaction={
        pattern=north west lines
    }] table[x=x,y=y] {
x y
1 0.344
2 0.344
3 0.344
};

\addplot [black,fill=RandomColor,postaction={
        pattern=crosshatch
    }] table[x=x,y=y] {
x y
1 0.394
2 0.387
3 0.365
    };

\node[anchor=south, align=left, rotate=45] at (axis cs: 1.2, 0.414) {\scriptsize +.050};
\node[anchor=south, align=left, rotate=45] at (axis cs: 2.2, 0.407) {\scriptsize +.043};
\node[anchor=south, align=left, rotate=45] at (axis cs: 3.2, 0.385) {\scriptsize +.021};

\end{axis}
    
\end{tikzpicture}
\vspace{-6mm}
\caption{Cache Dev Test Val}
\end{subfigure}
\caption{Ablation Study of Hierarchical Retrieval.} 
\label{fig:ablation}
\end{figure}
\input{figures/e2e_performance2}
\begin{figure}[t]
\usetikzlibrary{patterns}
\begin{subfigure}[b]{\linewidth}
\centering
\begin{tikzpicture}

\pgfplotstableread[col sep=comma,]{
name
10\%
30\%
50\%
70\%
90\%
}\datatable

\begin{axis}[
    ybar,
    clip=false,
    xtick={1, 2, 3, 4, 5, 6},
    xticklabels from table={\datatable}{name},
                 x tick label style={anchor=center, yshift = 0ex, font=\realfootnotesize},
    xtick style ={draw=none},
    xlabel style={yshift = 1.5ex, font=\realfootnotesize},
    ylabel style={yshift = -1ex, font=\realfootnotesize},
    width=80mm,
    height=32mm,
    bar width=2.0mm,
    ymin=0,
    ymax=0.4,
    log origin = infty,
    axis y line*=none,
    axis x line*=none,
    ytick={0, 0.1, 0.2, 0.3, 0.4},
    yticklabels={+0\%, +10\%, +20\%, +30\%, +40\%},
    xmin=0.5,
    xmax = 5.5,
    ymajorgrids,
    tick label style={font=\realfootnotesize},
    legend style={
        font=\realfootnotesize,
        /tikz/every even column/.append style={column sep=0.2cm},
        legend columns = 3,
        at={(-0.06,1.1)},
        anchor=south west,
    },
    xlabel={Apprentice VLM (Qwen) \%},
    ylabel={Qwen score +\%},
    area legend
    ]
    
\addplot [black,fill=NoOptColor, postaction={
        pattern=north west lines
}] table[x=x,y=y] {
x y
1 0.060
2 0.114
3 0.144
4 0.108
5 0.044
};
\addlegendentry[]{\nostart};
    
\addplot
    [black,fill=GreedyColor,postaction={
        pattern=crosshatch dots
    }] table[x=x,y=y] {
x y
1 0.213
2 0.177
3 0.153
4 0.153
5 0.141
    };
    \addlegendentry[]{\coldstart};

    \addplot [black,fill=BlueColor, postaction={
        pattern=bricks
    }] table[x=x,y=y] {
x y
1 0.366
2 0.188
3 0.213
4 0.223
5 0.258
    };
    \addlegendentry[]{\warmstart};

\end{axis}
    
\end{tikzpicture}
\end{subfigure}

\caption{Apprentice VLM MMMU score increases vs. Cache and Usage
}
\label{fig:experiment_checkout_undo}
\end{figure}
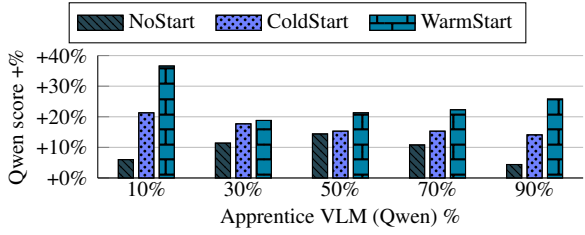
\begin{figure}[t]\captionsetup[subfigure]{font=footnotesize}
\pgfplotsset{scaled y ticks=false}
\usetikzlibrary{patterns}
\begin{subfigure}[b]{\linewidth}
\centering
\begin{tikzpicture}

\pgfplotstableread[col sep=comma,]{
name
10\%
30\%
50\%
70\%
90\%
}\datatable

\begin{axis}[
    ybar,
    clip=false,
    xtick={1, 2, 3, 4, 5, 6},
    xticklabels from table={\datatable}{name},
                 x tick label style={anchor=center, yshift = 0ex, font=\footnotesize},
    xtick style ={draw=none},
    xlabel style={yshift = 1.5ex, font=\realfootnotesize},
    ylabel style={yshift = -1ex, font=\realfootnotesize},
    width=80mm,
    height=32mm,
    bar width=2.0mm,
    ymin=-0.04,
    ymax=0.08,
    log origin = infty,
    axis y line*=none,
    axis x line*=none,
    ytick={-0.04, -0.02, 0, 0.02, 0.04, 0.06, 0.08},
    yticklabels={-0.04, -0.02, 0, 0.02, 0.04, 0.06, 0.08},
    xmin=0.5,
    xmax = 5.5,
    ymajorgrids,
    tick label style={font=\realfootnotesize},
    legend style={
        font=\realfootnotesize,
        /tikz/every even column/.append style={column sep=0.2cm},
        legend columns = 3,
        at={(-0.06,1.1)},
        anchor=south west,
    },
    xlabel={Apprentice VLM (Qwen) \%},
    ylabel={MMMU score inc.},
    area legend
    ]
    
\addplot [black,fill=NoOptColor,x tick label style={xshift=-0.3cm}, postaction={
        pattern=north west lines
}] table[x=x,y=y] {
x y
1 0.035
2 0.036
3 0.071
4 0.036
5 0.038
};
\addlegendentry[]{1st split};
    
\addplot
    [black,fill=GreedyColor,x tick label style={xshift=-0.3cm}, postaction={
        pattern=crosshatch dots
    }] table[x=x,y=y] {
x y

1 -0.031
2 -0.012
3 0.029
4 0.071
5 0.004
    };
    \addlegendentry[]{2nd split};

    \addplot [black,fill=BlueColor,x tick label style={xshift=-0.3cm}, postaction={
        pattern=bricks
    }] table[x=x,y=y] {
x y
1 0.069
2 0.085
3 0.050
4 0.036
5 0.004
    };
    \addlegendentry[]{3rd split};
\draw[black, very thick, opacity=0.5] (axis cs: 0.5, 0) -- (axis cs: 5.5, 0);
\end{axis}
    
\end{tikzpicture}
\end{subfigure}

\caption{Apprentice VLM MMMU score increases vs. Stage and Usage
}
\label{fig:experiment_checkout_three_stage}
\end{figure}
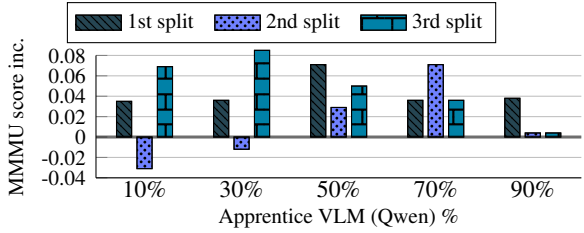

\section{Analysis}
\subsection{Static Performance}
This section preliminarily validates the effectiveness of CoT's multi-modal in-context learning in the static cache setting. We present our findings in Table~\ref{tab:Qwen}, which reports Qwen-7B's performance on the challenging MMMU dataset when enhanced by CoT's various multi-modal retrieval strategies. Tables~\ref{tab:Flamingo1} and \ref{tab:Flamingo2} report OpenFlamingo's performance on both MMMU and VL-ICL benchmarks when similarly enhanced by CoT. We experiment with two caching strategies on these datasets: storing a larger validation set (Val) while testing on a smaller development set (Dev), and vice versa.

\noindent \textbf{More Benefits with Larger Cache.}
As shown in Table~\ref{tab:Qwen}, caching a larger split while testing on a smaller split leads to greater performance gains when applying in-context learning on smaller models. A similar trend is observed when we manually restrict the cache size to half its original capacity.

\noindent \textbf{Hierarchical Retrieval Outperforms Dense Retrieval, but at a Cost.}
Our experiments aim to determine which retrieval method performs better and is more promising for in-context example selection. As shown in Table~\ref{tab:Qwen}, the hierarchical retrieval method slightly outperforms the dual-modality dense retrieval approach. However, this marginal advantage can only be achieved after fine-tuning a sensitive hyperparameter (cluster number). Figure~\ref{fig:ablation} details more related ablations.

\noindent \textbf{Best Dense Retrieval Strategy.}
We find that caching image embeddings, and querying with the average of image and text embeddings encoded from keywords extracted from VLM responses is the most effective dense retrieval strategy. Interestingly, (1) pure image retrieval, as reported in the OpenFlamingo paper \citep{flamingo}, also performs reasonably well, and (2) when the test set is relatively small (potentially introducing bias), pure image retrieval can even outperform other dense retrieval methods. In Table~\ref{tab:Qwen}, all presented text-based dense retrieval results utilize our LLM keyword extractor (\cref{sec:retrieval_method}), as the CLIP embedding context window is often too small to encompass complete VLM responses.

Based on these findings, we choose to evaluate our hierarchical retrieval method and the best-performing dense retrieval strategy in our following more important dynamic experiments (\cref{sec:experiment_dynamic}).

\noindent \textbf{Openflamingo-3B and 9B Receive Significant Benefits from CoT}. We present results for OpenFlamingo combined with the best dense retriever in Tables~\ref{tab:Flamingo1} and \ref{tab:Flamingo2}. CoT achieves high performance gains across the OpenFlamingo series, on both datasets, varying cache capacities, and the number of in-context examples. In MMMU, OpenFlamingo struggles to accommodate two-shot GPT-4o responses due to shorter context length, which limits the effectiveness of in-context learning.

\subsection{Dynamic Performance}
\label{sec:experiment_dynamic}
This section studies CoT's trade-offs between performance and cost advantage in the dynamic cache setting with various cache sizes. Figure \ref{fig:experiment_e2e} reports the MMMU score versus cache and retrieval configurations (within each sub-plot) and percentage of apprentice VLM usage (between each sub-plot). For \textit{NoStart}, \textit{ColdStart} and \textit{WarmStart}, we initialize CoT's cache as an empty set, the dev set, and the test set, respectively. The \textit{Hierarchical} setting uses the hierarchical retrieval method with CoT's cache startting with the dev set. Figure \ref{fig:experiment_checkout_undo} presents the performance gains of CoT's apprentice VLM versus different cache initialization methods. Figure \ref{fig:experiment_checkout_three_stage} shows the MMMU score increase of CoT's apprentice VLM on the first, second, and third (temporal, equal-sized) splits of the validation set. 

\noindent \textbf{Disparity Between Higher and Lower Use of Apprentice VLMs.} The less frequently apprentice VLMs are used, the better the overall performance. However, CoT mitigates this performance loss: as shown in Figure \ref{fig:experiment_e2e}, with \textit{WarmStart} and CoT enabled, the MMMU score with 90\% apprentice VLM usage matches the performance without CoT with only 70\% apprentice VLM usage.

\noindent \textbf{Performance gains from CoT in the dynamic setting.} CoT consistently provides performance gains in all settings, with higher gains in settings with higher ratio of apprentice VLM usage. Most notably in Figure \ref{fig:experiment_e2e}'s WarmStart setting, the MMMU score gain increased from 0.007 to 0.77 as the apprentice VLM usage increased sfrom 10\% to 90\%. 

\noindent \textbf{Hierarchical vs. Dense Retrieval.} Unlike in static settings, hierarchical cluster-based retrieval underperforms vs. dense retrieval, primarily due to its hyperparameters remaining fixed while the cache content evolves. We defer exploration of dynamic hyperparameter tuning to future work.

\noindent \textbf{Apprentice VLM performance gains.} CoT consistently achieves performance increases for the apprentice VLM no matter how much it has been used. As shown in Figure~\ref{fig:experiment_checkout_undo}, when the usage of apprentice VLM is between 30\% and 70\%, the performance gains are between 10\% and 23\%.

\noindent \textbf{Apprentice VLM performance gain vs. cache usage.} As CoT's cache grows as its multiplexer routes queries to the master VLM, the effectiveness of CoT's in-context learning will improve over time: as seen in Figure~\ref{fig:experiment_checkout_three_stage}), at 10\%-30\% apprentice VLM usage, the MMMU score increased between the first and third dataset splits, showing that the apprentice VLM was able to perform better as more entries were added to the CoT's cache. 

\section{Conclusion}
We proposed CoT, a VLM query serving framework that achieves an effective cost-quality trade-off for general reasoning. CoT interleaves large (master) and small (apprentice) VLM usage for query serving, caching high-quality responses generated from master VLMs to significantly boost the performance of apprentice VLMs via a novel multi-modal retrieval and in-context learning technique. CoT features a multiplexer to balance master VLM calls for cached example generation, and apprentice VLM calls, which uses a dual-modality dense retrieval mechanism to fetch the most relevant examples from the cache for in-context learning. We evaluate CoT on various challenging reasoning benchmarks and show that CoT's techniques can increase overall reasoning performance by up to 7.7\% under the same budget constraint, and specifically boosts the performance of apprentice VLMs by up to 36.6\%.

\subsection{Future Work} In the future, we plan to explore generalizing CoT's techniques to models handling other modalities (e.g., code, audio) where similar accuracy-cost trade-offs are present. Additionally, we aim to expand the application of CoT by integrating it with reinforcement learning for VLMs. This direction involves presenting and updating a memory space analogous to an external knowledge set, enabling VLMs to learn to retrieve through reinforcement learning, as early explored in VTool-R1~\cite{wu2025vtoolr1vlmslearnthink}. Furthermore, we intend to \textbf{conduct a more thorough comparison of in-context learning, supervised finetuning, and reinforcement learning approaches} in the context of VLM reasoning. 

\subsection{Reproducibility, AI usage and Artifact}
We have open-sourced our code to enable other researchers to reuse our inference framework with memory design. However, we do not guarantee that others will obtain exactly the same numbers as those reported in our paper, due to the inherent nondeterminism in LLM inference, as discussed in the blog~\cite{he2025nondeterminism}.

This work uses LLM to help polish writing.

This work is built upon open-source models and code, with adherence to all license terms and usage.

\section{Acknowledgement}
We sincerely thank Derek Hoiem for his insightful discussions and valuable feedback on our paper.

This work is supported by the National Science Foundation under grant numbers NSF 2217144, NSF 2106592, NSF 2433308, NSF 2229612 and NSF 2441601. This work used both the DeltaAI advanced computing and data resource, which is supported by the National Science Foundation (award OAC 2320345) and the State of Illinois, and the Delta advanced computing and data resource which is supported by the National Science Foundation (award OAC 2005572) and the State of Illinois. Delta and DeltaAI are joint efforts of the University of Illinois Urbana-Champaign and its National Center for Supercomputing Applications.

\section{Limitations}
First, we conduct extensive experiments in a setup featuring one master model and one apprentice model, and CoT also has potential to extend this approach to a multi-level master-apprentice framework (e.g., incorporating a 7B model, a 72B model, a 405B model, and a GPT4-o). Second, while our current focus is on image and text modalities, the framework can be extended to include additional modalities, such as video data, acoustic data, code data or other sensory inputs, given more multi-modal large models are ready to use.

\bibliography{custom}

\begin{thebibliography}{46}
\providecommand{\natexlab}[1]{#1}

\bibitem[{Aamodt and Plaza(1994)}]{cbr}
Agnar Aamodt and Enric Plaza. 1994.
\newblock Case-based reasoning: foundational issues, methodological variations, and system approaches.
\newblock \emph{AI Commun.}, 7(1):39–59.

\bibitem[{Alayrac et~al.(2022)Alayrac, Donahue, Luc, Miech, Barr, Hasson, Lenc, Mensch, Millican, Reynolds, Ring, Rutherford, Cabi, Han, Gong, Samangooei, Monteiro, Menick, Borgeaud, Brock, Nematzadeh, Sharifzadeh, Bi\'{n}kowski, Barreira, Vinyals, Zisserman, and Simonyan}]{flamingo}
Jean-Baptiste Alayrac, Jeff Donahue, Pauline Luc, Antoine Miech, Iain Barr, Yana Hasson, Karel Lenc, Arthur Mensch, Katherine Millican, Malcolm Reynolds, Roman Ring, Eliza Rutherford, Serkan Cabi, Tengda Han, Zhitao Gong, Sina Samangooei, Marianne Monteiro, Jacob~L Menick, Sebastian Borgeaud, Andy Brock, Aida Nematzadeh, Sahand Sharifzadeh, Miko\l~aj Bi\'{n}kowski, Ricardo Barreira, Oriol Vinyals, Andrew Zisserman, and Kar\'{e}n Simonyan. 2022.
\newblock \href {https://proceedings.neurips.cc/paper_files/paper/2022/file/960a172bc7fbf0177ccccbb411a7d800-Paper-Conference.pdf} {Flamingo: a visual language model for few-shot learning}.
\newblock In \emph{Advances in Neural Information Processing Systems}, volume~35, pages 23716--23736. Curran Associates, Inc.

\bibitem[{Anthropic(2024)}]{Sonnet}
Anthropic. 2024.
\newblock \href {https://www.anthropic.com/news/claude-3-5-sonnet} {Claude 3.5 sonnet}.
\newblock \url{https://www.anthropic.com/news/claude-3-5-sonnet}.
\newblock Accessed 25 Oct. 2024.

\bibitem[{Brohan et~al.(2023)Brohan, Brown, Carbajal, Chebotar, Chen, Choromanski, Ding, Driess, Dubey, Finn, Florence, Fu, Arenas, Gopalakrishnan, Han, Hausman, Herzog, Hsu, Ichter, Irpan, Joshi, Julian, Kalashnikov, Kuang, Leal, Lee, Lee, Levine, Lu, Michalewski, Mordatch, Pertsch, Rao, Reymann, Ryoo, Salazar, Sanketi, Sermanet, Singh, Singh, Soricut, Tran, Vanhoucke, Vuong, Wahid, Welker, Wohlhart, Wu, Xia, Xiao, Xu, Xu, Yu, and Zitkovich}]{rt2}
Anthony Brohan, Noah Brown, Justice Carbajal, Yevgen Chebotar, Xi~Chen, Krzysztof Choromanski, Tianli Ding, Danny Driess, Avinava Dubey, Chelsea Finn, Pete Florence, Chuyuan Fu, Montse~Gonzalez Arenas, Keerthana Gopalakrishnan, Kehang Han, Karol Hausman, Alex Herzog, Jasmine Hsu, Brian Ichter, Alex Irpan, Nikhil Joshi, Ryan Julian, Dmitry Kalashnikov, Yuheng Kuang, Isabel Leal, Lisa Lee, Tsang-Wei~Edward Lee, Sergey Levine, Yao Lu, Henryk Michalewski, Igor Mordatch, Karl Pertsch, Kanishka Rao, Krista Reymann, Michael Ryoo, Grecia Salazar, Pannag Sanketi, Pierre Sermanet, Jaspiar Singh, Anikait Singh, Radu Soricut, Huong Tran, Vincent Vanhoucke, Quan Vuong, Ayzaan Wahid, Stefan Welker, Paul Wohlhart, Jialin Wu, Fei Xia, Ted Xiao, Peng Xu, Sichun Xu, Tianhe Yu, and Brianna Zitkovich. 2023.
\newblock Rt-2: Vision-language-action models transfer web knowledge to robotic control.
\newblock In \emph{arXiv preprint arXiv:2307.15818}.

\bibitem[{Brown et~al.(2020)Brown, Mann, Ryder, Subbiah, Kaplan, Dhariwal, Neelakantan, Shyam, Sastry, Askell, Agarwal, Herbert-Voss, Krueger, Henighan, Child, Ramesh, Ziegler, Wu, Winter, Hesse, Chen, Sigler, Litwin, Gray, Chess, Clark, Berner, McCandlish, Radford, Sutskever, and Amodei}]{gpt3}
Tom Brown, Benjamin Mann, Nick Ryder, Melanie Subbiah, Jared~D Kaplan, Prafulla Dhariwal, Arvind Neelakantan, Pranav Shyam, Girish Sastry, Amanda Askell, Sandhini Agarwal, Ariel Herbert-Voss, Gretchen Krueger, Tom Henighan, Rewon Child, Aditya Ramesh, Daniel Ziegler, Jeffrey Wu, Clemens Winter, Chris Hesse, Mark Chen, Eric Sigler, Mateusz Litwin, Scott Gray, Benjamin Chess, Jack Clark, Christopher Berner, Sam McCandlish, Alec Radford, Ilya Sutskever, and Dario Amodei. 2020.
\newblock \href {https://proceedings.neurips.cc/paper_files/paper/2020/file/1457c0d6bfcb4967418bfb8ac142f64a-Paper.pdf} {Language models are few-shot learners}.
\newblock In \emph{Advances in Neural Information Processing Systems}, volume~33, pages 1877--1901. Curran Associates, Inc.

\bibitem[{Chen et~al.(2023)Chen, Hu, Luan, Sun, Changpinyo, Ritter, and Chang}]{info}
Yang Chen, Hexiang Hu, Yi~Luan, Haitian Sun, Soravit Changpinyo, Alan Ritter, and Ming-Wei Chang. 2023.
\newblock \href {https://doi.org/10.18653/v1/2023.emnlp-main.925} {Can pre-trained vision and language models answer visual information-seeking questions?}
\newblock In \emph{Proceedings of the 2023 Conference on Empirical Methods in Natural Language Processing}, pages 14948--14968, Singapore. Association for Computational Linguistics.

\bibitem[{Chen et~al.(2024)Chen, Xu, Qi, and Guo}]{ragvl}
Zhanpeng Chen, Chengjin Xu, Yiyan Qi, and Jian Guo. 2024.
\newblock \href {https://arxiv.org/abs/2407.21439} {Mllm is a strong reranker: Advancing multimodal retrieval-augmented generation via knowledge-enhanced reranking and noise-injected training}.
\newblock \emph{Preprint}, arXiv:2407.21439.

\bibitem[{Chu et~al.(2023)Chu, Qiao, Lin, Xu, Yang, Hu, Wei, Zhang, Zhang, Wei et~al.}]{mobilevlm}
Xiangxiang Chu, Limeng Qiao, Xinyang Lin, Shuang Xu, Yang Yang, Yiming Hu, Fei Wei, Xinyu Zhang, Bo~Zhang, Xiaolin Wei, et~al. 2023.
\newblock Mobilevlm: A fast, reproducible and strong vision language assistant for mobile devices.
\newblock \emph{arXiv preprint arXiv:2312.16886}.

\bibitem[{Chu et~al.(2024)Chu, Qiao, Zhang, Xu, Wei, Yang, Sun, Hu, Lin, Zhang et~al.}]{mobilevlm2}
Xiangxiang Chu, Limeng Qiao, Xinyu Zhang, Shuang Xu, Fei Wei, Yang Yang, Xiaofei Sun, Yiming Hu, Xinyang Lin, Bo~Zhang, et~al. 2024.
\newblock Mobilevlm v2: Faster and stronger baseline for vision language model.
\newblock \emph{arXiv preprint arXiv:2402.03766}.

\bibitem[{Ding et~al.(2024)Ding, Mallick, Wang, Sim, Mukherjee, R{\"u}hle, Lakshmanan, and Awadallah}]{hybridllm}
Dujian Ding, Ankur Mallick, Chi Wang, Robert Sim, Subhabrata Mukherjee, Victor R{\"u}hle, Laks V.~S. Lakshmanan, and Ahmed~Hassan Awadallah. 2024.
\newblock \href {https://openreview.net/forum?id=02f3mUtqnM} {Hybrid {LLM}: Cost-efficient and quality-aware query routing}.
\newblock In \emph{The Twelfth International Conference on Learning Representations}.

\bibitem[{Duan et~al.(2024)Duan, Yuan, Pumacay, Wang, Ehsani, Fox, and Krishna}]{manipulate}
Jiafei Duan, Wentao Yuan, Wilbert Pumacay, Yi~Ru Wang, Kiana Ehsani, Dieter Fox, and Ranjay Krishna. 2024.
\newblock Manipulate-anything: Automating real-world robots using vision-language models.
\newblock \emph{arXiv preprint arXiv:2406.18915}.

\bibitem[{Gao et~al.(2024{\natexlab{a}})Gao, Sarkar, Xia, Xiao, Wu, Ichter, Majumdar, and Sadigh}]{pgvlm2024}
Jensen Gao, Bidipta Sarkar, Fei Xia, Ted Xiao, Jiajun Wu, Brian Ichter, Anirudha Majumdar, and Dorsa Sadigh. 2024{\natexlab{a}}.
\newblock Physically grounded vision-language models for robotic manipulation.
\newblock In \emph{IEEE International Conference on Robotics and Automation (ICRA)}. IEEE.

\bibitem[{Gao et~al.(2024{\natexlab{b}})Gao, Xiong, Gao, Jia, Pan, Bi, Dai, Sun, Wang, and Wang}]{RAGsurvery}
Yunfan Gao, Yun Xiong, Xinyu Gao, Kangxiang Jia, Jinliu Pan, Yuxi Bi, Yi~Dai, Jiawei Sun, Meng Wang, and Haofen Wang. 2024{\natexlab{b}}.
\newblock \href {https://arxiv.org/abs/2312.10997} {Retrieval-augmented generation for large language models: A survey}.
\newblock \emph{Preprint}, arXiv:2312.10997.

\bibitem[{Gemini(2024)}]{gemini}
Team Gemini. 2024.
\newblock \href {https://arxiv.org/abs/2403.05530} {Gemini 1.5: Unlocking multimodal understanding across millions of tokens of context}.
\newblock \emph{Preprint}, arXiv:2403.05530.

\bibitem[{He and Lab(2025)}]{he2025nondeterminism}
Horace He and Thinking~Machines Lab. 2025.
\newblock \href {https://doi.org/10.64434/tml.20250910} {Defeating nondeterminism in llm inference}.
\newblock \emph{Thinking Machines Lab: Connectionism}.
\newblock Https://thinkingmachines.ai/blog/defeating-nondeterminism-in-llm-inference/.

\bibitem[{Hu et~al.(2021)Hu, Shen, Wallis, Allen-Zhu, Li, Wang, Wang, and Chen}]{lora}
Edward~J. Hu, Yelong Shen, Phillip Wallis, Zeyuan Allen-Zhu, Yuanzhi Li, Shean Wang, Lu~Wang, and Weizhu Chen. 2021.
\newblock \href {https://arxiv.org/abs/2106.09685} {Lora: Low-rank adaptation of large language models}.
\newblock \emph{Preprint}, arXiv:2106.09685.

\bibitem[{Hu et~al.(2022)Hu, Verga, Chen, Cohen, and Chen}]{murag}
Hexiang~(Frank) Hu, Pat Verga, Wenhu Chen, William~Weston Cohen, and Xi~Chen. 2022.
\newblock Murag: Multimodal retrieval-augmented generator.

\bibitem[{Jiang et~al.(2024)Jiang, Irvin, Wang, Chaudhry, Chen, and Ng}]{manyshot}
Yixing Jiang, Jeremy Irvin, Ji~Hun Wang, Muhammad~Ahmed Chaudhry, Jonathan~H. Chen, and Andrew~Y. Ng. 2024.
\newblock \href {https://arxiv.org/abs/2405.09798} {Many-shot in-context learning in multimodal foundation models}.
\newblock \emph{Preprint}, arXiv:2405.09798.

\bibitem[{Lewis et~al.(2020)Lewis, Perez, Piktus, Petroni, Karpukhin, Goyal, K\"{u}ttler, Lewis, Yih, Rockt\"{a}schel, Riedel, and Kiela}]{RAG}
Patrick Lewis, Ethan Perez, Aleksandra Piktus, Fabio Petroni, Vladimir Karpukhin, Naman Goyal, Heinrich K\"{u}ttler, Mike Lewis, Wen-tau Yih, Tim Rockt\"{a}schel, Sebastian Riedel, and Douwe Kiela. 2020.
\newblock \href {https://proceedings.neurips.cc/paper_files/paper/2020/file/6b493230205f780e1bc26945df7481e5-Paper.pdf} {Retrieval-augmented generation for knowledge-intensive nlp tasks}.
\newblock In \emph{Advances in Neural Information Processing Systems}, volume~33, pages 9459--9474. Curran Associates, Inc.

\bibitem[{Li et~al.(2025)Li, Huang, Ding, Park, and Chen}]{li2025sieve}
Zhaoheng Li, Silu Huang, Wei Ding, Yongjoo Park, and Jianjun Chen. 2025.
\newblock \href {https://doi.org/10.14778/3749646.3749725} {Sieve: Effective filtered vector search with collection of indexes}.
\newblock \emph{Proc. VLDB Endow.}, 18(11):4723–4736.

\bibitem[{Li et~al.(2023)Li, Pi, and Park}]{li2023sc}
Zhaoheng Li, Xinyu Pi, and Yongjoo Park. 2023.
\newblock \href {https://doi.org/10.1109/ICDE55515.2023.00393} {S/c: Speeding up data materialization with bounded memory}.
\newblock In \emph{2023 IEEE 39th International Conference on Data Engineering (ICDE)}, pages 1981--1994.

\bibitem[{Lin et~al.(2024{\natexlab{a}})Lin, Chen, Mei, Coca, and Byrne}]{finegrain}
Weizhe Lin, Jinghong Chen, Jingbiao Mei, Alexandru Coca, and Bill Byrne. 2024{\natexlab{a}}.
\newblock Fine-grained late-interaction multi-modal retrieval for retrieval augmented visual question answering.
\newblock In \emph{Proceedings of the 37th International Conference on Neural Information Processing Systems}, NIPS '23, Red Hook, NY, USA. Curran Associates Inc.

\bibitem[{Lin et~al.(2024{\natexlab{b}})Lin, Mei, Chen, and Byrne}]{lin-etal-2024-preflmr}
Weizhe Lin, Jingbiao Mei, Jinghong Chen, and Bill Byrne. 2024{\natexlab{b}}.
\newblock \href {https://aclanthology.org/2024.acl-long.289} {{P}re{FLMR}: Scaling up fine-grained late-interaction multi-modal retrievers}.
\newblock In \emph{Proceedings of the 62nd Annual Meeting of the Association for Computational Linguistics (Volume 1: Long Papers)}, pages 5294--5316, Bangkok, Thailand. Association for Computational Linguistics.

\bibitem[{Liu et~al.(2024)Liu, Wang, Sun, and Yu}]{rec}
Yuqing Liu, Yu~Wang, Lichao Sun, and Philip~S. Yu. 2024.
\newblock \href {https://arxiv.org/abs/2402.08670} {Rec-gpt4v: Multimodal recommendation with large vision-language models}.
\newblock \emph{Preprint}, arXiv:2402.08670.

\bibitem[{Malkov and Yashunin(2020)}]{HNSW}
Yu~A. Malkov and D.~A. Yashunin. 2020.
\newblock \href {https://doi.org/10.1109/TPAMI.2018.2889473} {Efficient and robust approximate nearest neighbor search using hierarchical navigable small world graphs}.
\newblock \emph{IEEE Trans. Pattern Anal. Mach. Intell.}, 42(4):824–836.

\bibitem[{Marino et~al.(2019)Marino, Rastegari, Farhadi, and Mottaghi}]{okvqa}
Kenneth Marino, Mohammad Rastegari, Ali Farhadi, and Roozbeh Mottaghi. 2019.
\newblock Ok-vqa: A visual question answering benchmark requiring external knowledge.
\newblock In \emph{Conference on Computer Vision and Pattern Recognition (CVPR)}.

\bibitem[{Mensink et~al.(2023)Mensink, Uijlings, Castrejon, Goel, Cadar, Zhou, Sha, Araujo, and Ferrari}]{evqa}
Thomas Mensink, Jasper Uijlings, Lluis Castrejon, Arushi Goel, Felipe Cadar, Howard Zhou, Fei Sha, Andre Araujo, and Vittorio Ferrari. 2023.
\newblock Encyclopedic {VQA}: Visual questions about detailed properties of fine-grained categories.
\newblock In \emph{ICCV}.

\bibitem[{Mialon et~al.(2023)Mialon, Dessi, Lomeli, Nalmpantis, Pasunuru, Raileanu, Roziere, Schick, Dwivedi-Yu, Celikyilmaz, Grave, LeCun, and Scialom}]{ragsv}
Gr{\'e}goire Mialon, Roberto Dessi, Maria Lomeli, Christoforos Nalmpantis, Ramakanth Pasunuru, Roberta Raileanu, Baptiste Roziere, Timo Schick, Jane Dwivedi-Yu, Asli Celikyilmaz, Edouard Grave, Yann LeCun, and Thomas Scialom. 2023.
\newblock \href {https://openreview.net/forum?id=jh7wH2AzKK} {Augmented language models: a survey}.
\newblock \emph{Transactions on Machine Learning Research}.
\newblock Survey Certification.

\bibitem[{Nguyen et~al.(2024)Nguyen, Liu, Li, Cai, Ojha, and Lee}]{assistant}
Thao Nguyen, Haotian Liu, Yuheng Li, Mu~Cai, Utkarsh Ojha, and Yong~Jae Lee. 2024.
\newblock \href {https://arxiv.org/abs/2406.09400} {Yo'llava: Your personalized language and vision assistant}.
\newblock \emph{Preprint}, arXiv:2406.09400.

\bibitem[{nmslib(2024)}]{hnswlib}
nmslib. 2024.
\newblock Hnswlib - fast approximate nearest neighbor search.
\newblock \url{https://github.com/nmslib/hnswlib}.

\bibitem[{OpenAI(2024{\natexlab{a}})}]{gpt4omini}
OpenAI. 2024{\natexlab{a}}.
\newblock \href {https://openai.com/index/gpt-4o-mini-advancing-cost-efficient-intelligence/} {Gpt-4o mini: advancing cost-efficient intelligence}.
\newblock \url{https://openai.com/index/gpt-4o-mini-advancing-cost-efficient-intelligence/}.
\newblock Accessed 25 Oct. 2024.

\bibitem[{OpenAI(2024{\natexlab{b}})}]{gpt4o}
OpenAI. 2024{\natexlab{b}}.
\newblock \href {https://openai.com/index/hello-gpt-4o/} {Hello gpt-4o}.
\newblock \url{https://openai.com/index/hello-gpt-4o/}.
\newblock Accessed 25 Oct. 2024.

\bibitem[{Pi et~al.(2024)Pi, Wu, Jiang, Zheng, Tian, Zhai, Nahrstedt, and Hu}]{uouo}
Xinyu Pi, Mingyuan Wu, Jize Jiang, Haozhen Zheng, Beitong Tian, ChengXiang Zhai, Klara Nahrstedt, and Zhiting Hu. 2024.
\newblock \href {https://aclanthology.org/2024.emnlp-main.369} {{UOUO}: Uncontextualized uncommon objects for measuring knowledge horizons of vision language models}.
\newblock In \emph{Proceedings of the 2024 Conference on Empirical Methods in Natural Language Processing}, pages 6432--6441, Miami, Florida, USA. Association for Computational Linguistics.

\bibitem[{Radford et~al.(2021)Radford, Kim, Hallacy, Ramesh, Goh, Agarwal, Sastry, Askell, Mishkin, Clark, Krueger, and Sutskever}]{CLIP}
Alec Radford, Jong~Wook Kim, Chris Hallacy, Aditya Ramesh, Gabriel Goh, Sandhini Agarwal, Girish Sastry, Amanda Askell, Pamela Mishkin, Jack Clark, Gretchen Krueger, and Ilya Sutskever. 2021.
\newblock \href {https://proceedings.mlr.press/v139/radford21a.html} {Learning transferable visual models from natural language supervision}.
\newblock In \emph{Proceedings of the 38th International Conference on Machine Learning}, volume 139 of \emph{Proceedings of Machine Learning Research}, pages 8748--8763. PMLR.

\bibitem[{Vendrow et~al.(2024)Vendrow, Pantazis, Shepard, Brostow, Jones, Mac~Aodha, Beery, and Van~Horn}]{inquire}
Edward Vendrow, Omiros Pantazis, Alexander Shepard, Gabriel Brostow, Kate~E Jones, Oisin Mac~Aodha, Sara Beery, and Grant Van~Horn. 2024.
\newblock Inquire: A natural world text-to-image retrieval benchmark.
\newblock \emph{NeurIPS}.

\bibitem[{Wang et~al.(2024)Wang, Bai, Tan, Wang, Fan, Bai, Chen, Liu, Wang, Ge, Fan, Dang, Du, Ren, Men, Liu, Zhou, Zhou, and Lin}]{Qwen2VL}
Peng Wang, Shuai Bai, Sinan Tan, Shijie Wang, Zhihao Fan, Jinze Bai, Keqin Chen, Xuejing Liu, Jialin Wang, Wenbin Ge, Yang Fan, Kai Dang, Mengfei Du, Xuancheng Ren, Rui Men, Dayiheng Liu, Chang Zhou, Jingren Zhou, and Junyang Lin. 2024.
\newblock Qwen2-vl: Enhancing vision-language model's perception of the world at any resolution.
\newblock \emph{arXiv preprint arXiv:2409.12191}.

\bibitem[{Wang et~al.(2018)Wang, Wu, Shen, Dick, and van~den Hengel}]{FVQA}
Peng Wang, Qi~Wu, Chunhua Shen, Anthony Dick, and Anton van~den Hengel. 2018.
\newblock \href {https://doi.org/10.1109/TPAMI.2017.2754246} {Fvqa: Fact-based visual question answering}.
\newblock \emph{IEEE Transactions on Pattern Analysis and Machine Intelligence}, 40(10):2413--2427.

\bibitem[{Wu et~al.(2025)Wu, Yang, Jiang, Li, Yan, Yu, Zhang, Zhai, and Nahrstedt}]{wu2025vtoolr1vlmslearnthink}
Mingyuan Wu, Jingcheng Yang, Jize Jiang, Meitang Li, Kaizhuo Yan, Hanchao Yu, Minjia Zhang, Chengxiang Zhai, and Klara Nahrstedt. 2025.
\newblock \href {https://arxiv.org/abs/2505.19255} {Vtool-r1: Vlms learn to think with images via reinforcement learning on multimodal tool use}.
\newblock \emph{Preprint}, arXiv:2505.19255.

\bibitem[{Yasunaga et~al.(2023)Yasunaga, Aghajanyan, Shi, James, Leskovec, Liang, Lewis, Zettlemoyer, and Yih}]{RACM3}
Michihiro Yasunaga, Armen Aghajanyan, Weijia Shi, Richard James, Jure Leskovec, Percy Liang, Mike Lewis, Luke Zettlemoyer, and Wen-Tau Yih. 2023.
\newblock \href {https://proceedings.mlr.press/v202/yasunaga23a.html} {Retrieval-augmented multimodal language modeling}.
\newblock In \emph{Proceedings of the 40th International Conference on Machine Learning}, volume 202 of \emph{Proceedings of Machine Learning Research}, pages 39755--39769. PMLR.

\bibitem[{Yuan et~al.(2024)Yuan, Sun, Omeiza, Zhao, Newman, Kunze, and Gadd}]{RAG-drive}
Jianhao Yuan, Shuyang Sun, Daniel Omeiza, Bo~Zhao, Paul Newman, Lars Kunze, and Matthew Gadd. 2024.
\newblock {RAG-Driver: Generalisable Driving Explanations with Retrieval-Augmented In-Context Learning in Multi-Modal Large Language Model}.
\newblock In \emph{Proceedings of Robotics: Science and Systems}, Delft, Netherlands.

\bibitem[{Yue et~al.(2024)Yue, Ni, Zhang, Zheng, Liu, Zhang, Stevens, Jiang, Ren, Sun, Wei, Yu, Yuan, Sun, Yin, Zheng, Yang, Liu, Huang, Sun, Su, and Chen}]{mmmu}
Xiang Yue, Yuansheng Ni, Kai Zhang, Tianyu Zheng, Ruoqi Liu, Ge~Zhang, Samuel Stevens, Dongfu Jiang, Weiming Ren, Yuxuan Sun, Cong Wei, Botao Yu, Ruibin Yuan, Renliang Sun, Ming Yin, Boyuan Zheng, Zhenzhu Yang, Yibo Liu, Wenhao Huang, Huan Sun, Yu~Su, and Wenhu Chen. 2024.
\newblock Mmmu: A massive multi-discipline multimodal understanding and reasoning benchmark for expert agi.
\newblock In \emph{Proceedings of CVPR}.

\bibitem[{Zhang et~al.(2025)Zhang, Zhang, Dong, Zang, and Wang}]{longclip}
Beichen Zhang, Pan Zhang, Xiaoyi Dong, Yuhang Zang, and Jiaqi Wang. 2025.
\newblock Long-clip: Unlocking the long-text capability of clip.
\newblock In \emph{Computer Vision -- ECCV 2024}, pages 310--325, Cham. Springer Nature Switzerland.

\bibitem[{Zhang et~al.(2023)Zhang, Zhou, and Liu}]{visualcontext}
Yuanhan Zhang, Kaiyang Zhou, and Ziwei Liu. 2023.
\newblock \href {https://proceedings.neurips.cc/paper_files/paper/2023/file/398ae57ed4fda79d0781c65c926d667b-Paper-Conference.pdf} {What makes good examples for visual in-context learning?}
\newblock In \emph{Advances in Neural Information Processing Systems}, volume~36, pages 17773--17794. Curran Associates, Inc.

\bibitem[{Zhang et~al.(2024)Zhang, Zhang, Ding, and Yue}]{search}
Zhixin Zhang, Yiyuan Zhang, Xiaohan Ding, and Xiangyu Yue. 2024.
\newblock Vision search assistant: Empower vision-language models as multimodal search engines.
\newblock \emph{arXiv preprint arXiv:2410.21220}.

\bibitem[{Zhou et~al.(2024)Zhou, Liu, Yurtsever, Zagar, Zimmer, Cao, and Knoll}]{ad}
Xingcheng Zhou, Mingyu Liu, Ekim Yurtsever, Bare~Luka Zagar, Walter Zimmer, Hu~Cao, and Alois~C. Knoll. 2024.
\newblock \href {https://doi.org/10.1109/TIV.2024.3402136} {Vision language models in autonomous driving: A survey and outlook}.
\newblock \emph{IEEE Transactions on Intelligent Vehicles}, pages 1--20.

\bibitem[{Zong et~al.(2024)Zong, Bohdal, and Hospedales}]{vl-icl}
Yongshuo Zong, Ondrej Bohdal, and Timothy Hospedales. 2024.
\newblock \href {https://arxiv.org/abs/2403.13164} {Vl-icl bench: The devil in the details of multimodal in-context learning}.
\newblock \emph{Preprint}, arXiv:2403.13164.

\end{thebibliography}
\newpage
\appendix
\onecolumn
\section{Appendix}
\label{sec:appendix}
\begin{tcolorbox}[
    colback=red!5!white,
    colframe=red!75!black,
    title={\textbf{Open-Flamingo Prompts}},
    fonttitle=\bfseries
]

\textbf{MMMU In-context Learning Prompt:}\\
{\color{blue}\textnormal{\texttt{\color{teal}<image>} \texttt{\color{blue} <shot-1>} \texttt{\color{teal}<|endofchunk|>} ... 
\texttt{\color{teal}<image>} \texttt{\color{blue} <shot-n>} \texttt{\color{teal}<|endofchunk|> \texttt{\color{blue}<question>}} 
\texttt{\color{teal}<image>}}}
The options are the following: {\color{blue} \textnormal{\texttt{<option>}}}
There is only one option possible. The answer is

\bigskip

\textbf{MMMU N-shot Cache Sample:}\\
{\color{red} Human:} \textnormal{\texttt{\color{blue}<question>} The options are the following: \texttt{\color{blue}<option>}}.
Please include your reasoning steps, then answer your choice in this format:
ANSWER: LETTER CHOICE.
The letter choice is strictly in alphabetical order, and there is only one option possible.

{\color{blue} Assistant:} \textnormal{\texttt{\color{blue}<gpt4-o response>}}

\bigskip

\textbf{MMMU Auxiliary VLM Prompt for Response:} \\
{\color{blue} \textnormal{\texttt{\color{teal}<image>} \texttt{<question>}}}
The options are the following: {\color{blue} \textnormal{\texttt{<option>}}}
There is only one option possible. The answer is

\bigskip

\textbf{CLEVR In-context Learning Prompt:}\\
{\color{blue} \textnormal{\texttt{\color{teal}<image>} \texttt{<shot-1>} \texttt{\color{teal}<|endofchunk|>} ... 
\texttt{\color{teal}<image>} \texttt{<shot-n>} \texttt{\color{teal}<|endofchunk|>} 
\texttt{\color{teal}<image>}}}
How many {\color{blue} \textnormal{\texttt{<question>}}} objects? The answer is

\bigskip

\textbf{CLEVR N-shot Cache Sample:}\\
{\color{red} Human:} How many objects in the image have the
{\color{blue} \textnormal{\texttt{<question>}}}?
Please include your reasoning steps, then answer your choice in this format:
ANSWER: NUMBER.

{\color{blue} Assistant:} \textnormal{\color{blue}\texttt{<gpt4-o response>}}

\bigskip

\textbf{CLEVR Auxiliary VLM Prompt for Response:}\\
{\color{teal} \textnormal{\texttt{<image>}}} How many {\color{blue} \textnormal{\texttt{<question>}}} objects?
The answer is

\bigskip

\textbf{TextOCR In-context Learning Prompt:}\\
{\color{blue} \textnormal{\texttt{\color{teal}<image>} \texttt{<shot-1>} \texttt{\color{teal}<|endofchunk|>} ...
\texttt{\color{teal}<image>} \texttt{<shot-n>} \texttt{\color{teal}<|endofchunk|>} 
\texttt{\color{teal}<image>}}}
What text is shown in the red box? Only answer with the largest text. The answer is

\bigskip

\textbf{TextOCR N-shot Cache Sample:}\\
{\color{red} Human:} What text is shown in the red box?
Only answer with the largest text. Please include your reasoning steps, then answer your choice in this format:
ANSWER: TEXT.

{\color{blue} Assistant:} \textnormal{\color{blue}\texttt{<gpt4-o response>}}

\bigskip

\textbf{TextOCR Auxiliary VLM Prompt for Response:}\\
{\color{blue} \textnormal{\texttt{\color{teal}<image>}}} What text is shown in the red box?
Only answer with the largest text. The answer is

\bigskip

\end{tcolorbox}

\begin{tcolorbox}[
    breakable,
    colback=red!5!white,
    colframe=red!75!black,
    title={\textbf{Qwen Prompts}},
    fonttitle=\bfseries
]

\textbf{MMMU In-context Learning Prompt:}\\
This is a chat between a curious human and an artificial intelligence assistant. The assistant gives helpful, detailed, and polite answers to the human's questions. If a question does not make any sense, or is not factually coherent, explain why instead of answering something not correct. If you don't know the answer to a question, please don't share false information. The assistant will have one similar in-context example provided by another powerful assistant:

\texttt{\color{blue}<shot-1>...<shot-n>}

Now you should answer the following question given the image below and you can use GPT4 Assistant's case for reference:\\
{\color{red} Human:} \\
 \texttt{\color{blue}<question> <option>} Please include your reasoning steps, then answer your choice in this format: ANSWER: \texttt{<LETTER CHOICE>}. The letter choice is strictly in the alphabetical order, and there is only one option possible.\\
{\color{blue} Assistant(you):}

\bigskip
\textbf{MMMU N-shot Cache Sample:}\\
{\color{red}Human:} \texttt{\color{blue}<question> <option>} Please include your reasoning steps, then answer your choice in this format: ANSWER: \texttt{<LETTER CHOICE>}. The letter choice is strictly in the alphabetical order, and there is only one option possible.\\
{\color{blue}Assistant:} \texttt{\color{blue}<gpt4-o response>}

\bigskip
\textbf{MMMU Auxiliary VLM Prompt for Response:} \\
Now you should answer the following question:\\
{\color{red} Human:} \\
\texttt{\color{blue}<question> <option>} Please include your reasoning steps, then answer your choice in this format: ANSWER: \texttt{<LETTER CHOICE>}. The letter choice is strictly in the alphabetical order, and there is only one option possible.\\
{\color{blue} Assistant(you):}
\end{tcolorbox}

\begin{tcolorbox}[
    breakable,
    colback=red!5!white,
    colframe=red!75!black,
    title={\textbf{Llama Auxiliary Keyword Extractor Prompts}},
    fonttitle=\bfseries
]
\textbf{\textcolor{black}{MMMU Prompt}}\\
You are a helpful chatbot that assists users in generating keywords from conversations. You have been given a piece of text and need to generate keywords from it. Please give me 10 keywords that are present in this question-option context and separate them with commas. Make sure you to only return the keywords and say nothing else. Make sure to exclude the words "question" and "options" as keywords I have the following multiple choice question and its options: \texttt{\color{blue}<query>}

\bigskip
\textbf{\textcolor{black}{CLEVR and TextOCR Prompt}}\\
You are a helpful chatbot that assists users in generating keywords from conversations. You have been given a piece of text and need to generate keywords from it. 
Please give me 10 keywords that are present in this context and separate them with commas.
Make sure you to only return the keywords and say nothing else.
Make sure to exclude the word "ANSWER" as keywords
I have the following contexts: \texttt{\color{blue}<query>}

\end{tcolorbox}

\begin{tcolorbox}[
    breakable,
    colback=red!5!white,
    colframe=red!75!black,
    title={\textbf{GPT4-o Prompts}},
    fonttitle=\bfseries
]
\textbf{\textcolor{black}{MMMU Prompt}}\\
\textnormal{\texttt{\color{blue}<question>} The options are the following: \texttt{\color{blue}<option>}}.
Please include your reasoning steps, then answer your choice in this format:
ANSWER: \texttt{<LETTER CHOICE>}.
The letter choice is strictly in alphabetical order, and there is only one option possible.

\bigskip
\textbf{\textcolor{black}{CLEVR Prompt}}\\
How many objects in the image have the {\color{blue} \textnormal{\texttt{<question>}}} Please include your reasoning steps, then answer your choice in this format: ANSWER: \texttt{<NUMBER>}.

\bigskip
\textbf{\textcolor{black}{TextOCR Prompt}}\\
What text is shown in the red box? Only answer with the largest text. Please include your reasoning steps, then answer your choice in this format: ANSWER: \texttt{<TEXT>}.
\end{tcolorbox}

\end{document}